%% file: acl_latex.tex
\pdfoutput=1
\documentclass[11pt]{article}

\usepackage[final]{acl}

\usepackage{times}
\usepackage{latexsym}

\usepackage[T1]{fontenc}

\usepackage[utf8]{inputenc}

\usepackage{microtype}

\usepackage{inconsolata}

\usepackage{amssymb}%
\usepackage{pifont}%
\usepackage{graphicx}
\usepackage{hyperref}
\usepackage{url}
\usepackage{microtype}
\usepackage{times}
\usepackage{latexsym}
\usepackage{soul}
\usepackage{amsmath}
\usepackage{array, booktabs}
\usepackage{bm}
\usepackage{cleveref}
\usepackage{graphicx}
\usepackage{tabularx}
\usepackage{soul}
\usepackage{xcolor}
\usepackage{todonotes}
\usepackage{multirow}
\usepackage{xpatch}
\usepackage{blindtext}
\usepackage{fdsymbol}
\usepackage{microtype}
\usepackage{hyperref}
\usepackage[export]{adjustbox}
\usepackage{textcomp}
\usepackage{xparse}
\usepackage{enumitem}
\usepackage{makecell}
\usepackage{subcaption}
\usepackage{colortbl}
\usepackage{tcolorbox}
\usepackage{xparse}
\usepackage{stfloats}
\usepackage{arydshln}
\usepackage{bbm}

\usepackage[linesnumbered,vlined,ruled]{algorithm2e}

\input{content/commands}

\title{Why Vision Language Models Struggle with Visual Arithmetic?\\ Towards Enhanced Chart and Geometry Understanding}

\author{Kung-Hsiang Huang$^{1}$ ~~~Can Qin$^{1}$ ~~~Haoyi Qiu$^{2}$ \\ {\bfseries Philippe Laban$^{1}$ ~~~Shafiq Joty$^{1}$ ~~~Caiming Xiong$^{1}$ ~~~Chien-Sheng Wu$^{1}$} \\
$^{1}$Salesforce AI Research ~~ $^{2}$UCLA\\
$^{1}$\texttt{\{kh.huang, cqin, sjoty, cxiong, wu.jason\}@salesforce.com} \\
}

\begin{document}
\maketitle
\input{content/00_abstract}

\input{content/01_introduction}

\input{content/03_probing}
\input{content/04_method}

\input{content/05_results}

\input{content/02_related_work}
\input{content/06_conclusion}

\input{content/07_limitation}

\bibliography{custom}

\input{content/appendix}

\end{document}

%% file: content/commands.tex
\newcolumntype{P}[1]{>{\centering\arraybackslash}p{#1}}

\newcommand{\method}[1]{\textsc{CogAlign}}
\newcommand{\data}[1]{\textsc{CogAlign64k}}

\newcommand{\cmark}{\ding{51}}%
\newcommand{\xmark}{\ding{55}}%
\definecolor{c2}{RGB}{218,0,0}

\definecolor{lightblue}{RGB}{212, 235, 255}
\definecolor{lightorange}{RGB}{255, 204, 168}
\definecolor{lightyellow}{RGB}{255, 255, 168}
\definecolor{lightred}{RGB}{255, 168, 168}
\definecolor{darkred}{RGB}{234, 107, 102}
\definecolor{darkerblue}{RGB}{103, 136, 184}
\definecolor{lightgreen}{RGB}{102, 205, 102}

\definecolor{gold}{rgb}{0.83, 0.69, 0.22}
\sethlcolor{lightblue}
\newcolumntype{Y}{>{\centering\arraybackslash}X}

\NewDocumentCommand{\steeve}
{ mO{} }{\textcolor{gold}{\textsuperscript{\textit{Steeve}}\textsf{\textbf{\small[#1]}}}}
\NewDocumentCommand{\jason}
{ mO{} }{\textcolor{red}{\textsuperscript{\textit{Jason}}\textsf{\textbf{\small[#1]}}}}

\NewDocumentCommand{\shafiq}
{ mO{} }{\textcolor{blue}{\textsuperscript{\textit{Shafiq}}\textsf{\textbf{\small[#1]}}}}
\NewDocumentCommand{\sid}
{ mO{} }{\textcolor{green}{\textsuperscript{\textit{Sid}}\textsf{\textbf{\small[#1]}}}}
\NewDocumentCommand{\philippe}
{ mO{} }{\textcolor{magenta}{\textsuperscript{\textit{Philippe}}\textsf{\textbf{\small[#1]}}}}
\NewDocumentCommand{\can}
{ mO{} }{\textcolor{violet}{\textsuperscript{\textit{Can}}\textsf{\textbf{\small[#1]}}}}
\NewDocumentCommand{\caiming}
{ mO{} }{\textcolor{orange}{\textsuperscript{\textit{Caiming}}\textsf{\textbf{\small[#1]}}}}
\NewDocumentCommand{\ap}
{ mO{} }{\textcolor{brown}{\textsuperscript{\textit{Akshara}}\textsf{\textbf{\small[#1]}}}}

\definecolor{gold}{rgb}{0.83, 0.69, 0.22}

\newcommand{\markgreen}[1]{{\color{lightgreen}#1}}

\newcommand{\mathv}[1]{\textsc{Math-Vision}}
\newcommand{\chocolate}[1]{\textsc{Chocolate}}

\crefformat{section}{\S#2#1#3} %
\crefformat{subsection}{\S#2#1#3}
\crefformat{subsubsection}{\S#2#1#3}

%% file: content/00_abstract.tex
\begin{abstract}

Vision Language Models (VLMs) have achieved remarkable progress in multimodal tasks, yet they often struggle with visual arithmetic, seemingly simple capabilities like object counting or length comparison, %
which are essential for relevant complex tasks like chart understanding and geometric reasoning. In this work, we first investigate the root causes of this deficiency through a suite of probing tasks focusing on basic visual arithmetic. Our analysis reveals that while pre-trained vision encoders typically capture sufficient information, the text decoder often fails to decode it correctly for arithmetic reasoning. To address this, we propose \method~, a novel post-training strategy inspired by Piaget's theory of cognitive development.  \method~ trains VLMs to recognize invariant properties under visual transformations. %
We demonstrate that this approach significantly improves the performance of three diverse VLMs on our proposed probing tasks. Furthermore, \method~ enhances performance by an average of 4.6\% on \chocolate~ and 2.9\% on \mathv~, outperforming or matching supervised fine-tuning methods while requiring only 60\% less training data. These results highlight the effectiveness and generalizability of \method~ in improving fundamental visual arithmetic capabilities and their transfer to downstream tasks. \footnote{\method~ data has been released at: \url{https://github.com/SalesforceAIResearch/CogAlign}.}

\end{abstract}

%% file: content/01_introduction.tex
\section{Introduction}

In recent years, vision language models (VLMs) have rapidly advanced, demonstrating remarkable capabilities in integrating and processing multimodal information \cite{liu2023llava, dai2023instructblip, chen2024internvl, xue2024xgen}. These models have found extensive applications across various domains, ranging from visual commonsense reasoning to sophisticated tasks like web agents \cite{xu2024llavacot, zhang2024improve, xie2024osworld, lin2024showui}. By leveraging both visual and textual data, VLMs promise a nuanced understanding that surpasses what can be achieved by analyzing them individually. %

Despite these advancements, current VLMs exhibit noticeable deficiencies in performing fundamental \textit{visual arithmetic}: these models struggle with seemingly simple tasks like accurately counting objects, comparing lengths, assessing angles, and evaluating relative sizes or areas \cite{Rahmanzadehgervi_2024_blind, wang2024vdlm, huang2024frompixels, ullman2024illusion, wei2024slow, kamoi2024visonlyqa}. These shortcomings are particularly evident in complex tasks such as chart understanding \cite{huang-etal-2024-lvlms} and geometric problem-solving \cite{gao2023gllava}. \looseness=-1%

In this study, we first delve into the root causes of VLMs' difficulties with visual arithmetic, exploring several hypotheses to elucidate why VLMs often fail when faced with such challenges (\Cref{sec:probing}). We propose a suite of probing tasks, focusing on basic visual arithmetic such as length comparison, to answer this question. %
Our analysis reveals that pre-trained vision encoders coupled with a simple linear classifier perform poorly on these probing tasks, indicating that a single linear layer is insufficient to decode the complex visual representations for arithmetic reasoning. However, when we fine-tune the  text decoder of a VLM on these tasks, performance significantly improves.  This suggests \textbf{the bottleneck lies in the decoder's ability to effectively process and utilize the visual information, rather than in the visual representation itself}.

To tackle these challenges, we propose a novel post-training strategy, \method~, designed to improve the performance of VLMs in visual arithmetic tasks (\Cref{sec:method}). Drawing inspiration from Piaget's theory of cognitive development \cite{piaget1952origins}, our method focuses on enhancing VLMs' understanding of \textit{conservation} (recognizing that certain properties remain unchanged despite transformations) and \textit{decentration} (considering multiple aspects simultaneously). We train VLMs using synthetically generated image pairs that demonstrate transformations, enabling them to compare and evaluate based on specific properties like length, angle, and quantity. By employing Direct Preference Optimization (DPO) \cite{rafailov2023dpo}, the model learns from both positive and negative examples, offering a richer learning signal than traditional Supervised Fine-Tuning (SFT). Our experiments show that \method~ significantly enhances performance across three VLMs of different scales and architectures on the proposed probing tasks.

\begin{figure*}[t]
    \centering
    \includegraphics[width=0.9\linewidth, trim=0 0 0 15, clip]{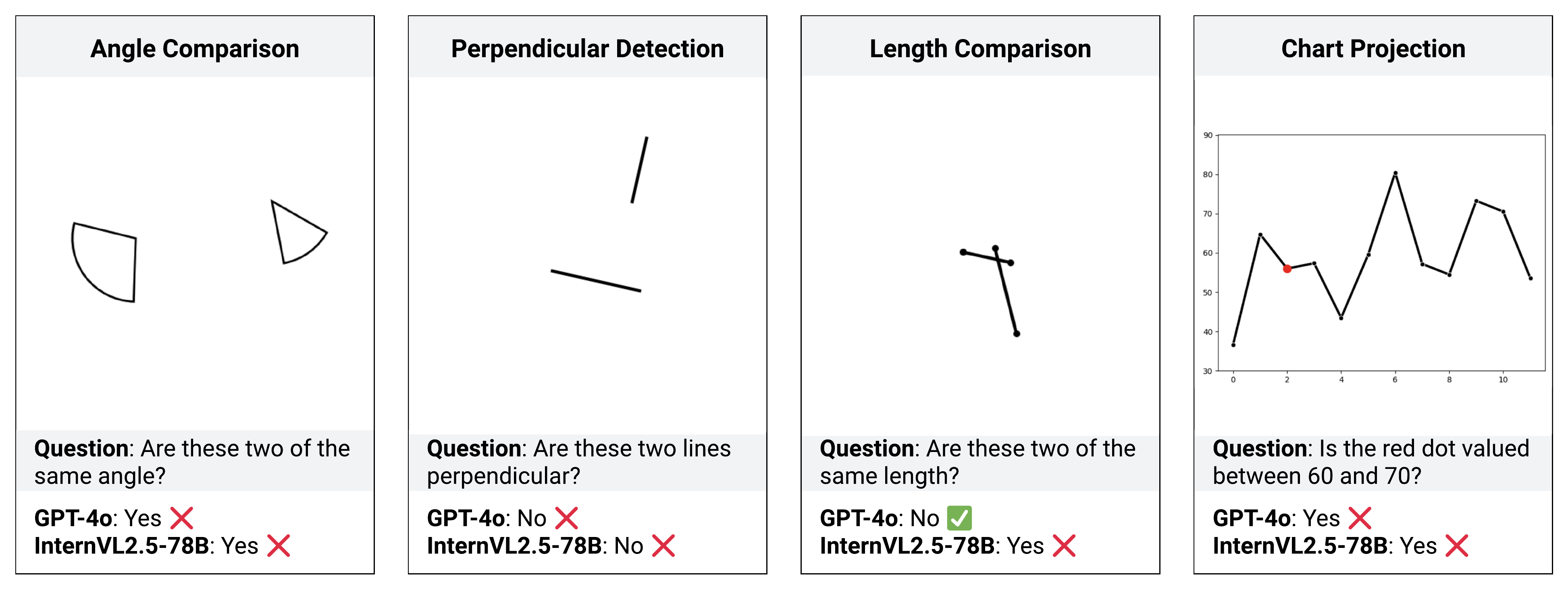}
    \vspace{-4mm}
    \caption{Examples of probing tasks designed to assess visual arithmetic abilities. Each task presents a visual input and a question requiring comparison or evaluation of geometric properties. At the bottom of each task, we see that even top-performing VLMs like GPT-4o and InternVL2.5-78B struggle with these seemly simple tasks.} %
    \vspace{-5mm}
    \label{fig:probing_tasks}
\end{figure*}

Furthermore, we evaluate \method~ on two downstream benchmarks: \chocolate~ \cite{huang-etal-2024-lvlms} for chart understanding, and \mathv~ \cite{wang2024mathv} for geometric problem-solving (\Cref{sec:exps}). Our results demonstrate the effectiveness of \method~ in enhancing performance on these complex tasks. On average, \method~ %
boosts performance by 4.6\% and 2.9\% on \chocolate~ and \mathv~ respectively, demonstrating that improving fundamental visual arithmetic capabilities translates to improved performance on downstream tasks. Notably, \method~ outperforms or achieves comparable performance to SFT methods while requiring 60\% less training data, even though \method~ does not involve direct optimization for specific tasks. %
This showcases its strong generalizability and highlights its potential of focusing on foundational skills to unlock broader capabilities in VLMs. \looseness=-1

Our main contributions are as follows:
\begin{itemize}[leftmargin=*]\itemsep0em 
    \item We conduct an in-depth analysis to uncover the root causes of VLMs' underperformance in tasks that involve visual arithmetic. \looseness=-1
    \item We develop \method~, a post-training strategy designed to enhance VLMs' abilities in understanding performing visual arithmetic.
    \item Extensive experiments on three VLMs show that \method~ significantly improves performance in chart comprehension and geometric problem-solving, highlighting its generalizability.
\end{itemize}

%% file: content/03_probing.tex
\section{Why Vision Language Models Struggle with Visual Arithmetic?}
\label{sec:probing}
As suggested in previous studies, VLMs struggle with visual arithmetic \cite{Rahmanzadehgervi_2024_blind, wang2024vdlm}, leading to poor performance in tasks involving such capabilities such as chart understanding \cite{huang-etal-2024-lvlms} and geometric problem-solving \cite{gao2023gllava}. In this section, we aim to understand the root causes behind such phenomenon. We first propose a suite of probing tasks we design to facilitate our analysis (\Cref{subsec:probing_tasks}) and then illustrate the various analyses we conduct to validate our hypotheses (\Cref{subsec:probing_analysis}).

\input{tables/probing_mlp}

\subsection{Probing Tasks}
\label{subsec:probing_tasks}

We propose four probing tasks for assessing visual arithmetic capabilities, motivated by the fundamental operations needed to interpret visual data quantitatively. For a VLM to successfully understand a chart, for example, it must be able to compare lengths of bars or lines, discern relationships indicated by line slopes, and projecting points onto axes. An overview of the probing tasks are shown in \Cref{fig:probing_tasks}. All four tasks are discriminative and can be considered binary classification tasks.  %
Below, we illustrate these tasks in details.

\paragraph{Angle Comparison} asks models to determine whether the angle of two wedges are the same. This requires the model to differentiate and measure angular magnitude, a seemly more complex operation that tests the model's grasp of spatial relationships and angular geometry. This task assesses the model's capacity to interpret rotational dimensions and engage in deeper analytical processing to distinguish subtle differences in angle, thereby evaluating the core geometric understanding of the model in angular perception. %
\vspace{-2mm}
\paragraph{Perpendicular Detection} challenges models to determine if two given lines are perpendicular to each other. Building upon the concept of angles, this task requires a deeper understanding of specific angular relationships, where perpendicularity implies a $90^\circ$ angle.  While Angle Comparison focuses on general angle differentiation, Perpendicular Detection assesses a model's ability to recognize this specific geometric configuration. 
\vspace{-2mm}

\paragraph{Length Comparison} asks models whether two lines with arbitrary slopes are of the same lengths. In addition to basic spatial reasoning, this task requires models to consider trigonometric relationships between the lines, demanding higher-level understanding of equivalence regardless of orientation. The variability in slopes necessitates an advanced ability to rotate or translate lines, challenging the model's proficiency in geometric reasoning beyond simple horizontal and vertical comparisons.

\vspace{-2mm}
\paragraph{Chart Projection} challenges the model to determine if the value of a red dot on a black line chart lies between 60 and 70. As the most complex task, this task integrates key aspects of the preceding tasks. It requires spatial reasoning to project the dot's position onto the y-axis, similar to Angle and Perpendicular Detection. It then involves comparing the projected value's magnitude against the specified range, akin to Length Comparison. %

\subsection{Probing Analysis}
\label{subsec:probing_analysis}

The research question we aim to answer is: \textbf{\textit{Do visual representations from pre-trained vision encoders contain enough information to perform visual arithmetic tasks?}} To answer this question, we conduct experiments by feeding the outputs from various encoders into a linear classifier to perform binary classification on the probing tasks. For each task, we randomly generate 12,000 images programmatically with a train:development:test split of 10:1:1. Each split has a balanced portion of positive and negative labels. We test a wide range of vision encoder, including  CLIP ViT-L/14 \cite{Radford2021LearningTV}, SigLIP-SO400M/14 \cite{Zhai_2023_siglip}, InternViT-300M-V2.5 \cite{chen2024internvl}, and DINOv2-Large \cite{oquab2024dinov}.%
We also evaluate the features produced by the projection layer of LLaVA-v1.5 \cite{liu2023llava}. Each model (i.e., the single classifier) %
was trained for 200 epochs and the checkpoint that achieves the highest performance on the development set is selected. \looseness=-1

The results are presented in \Cref{tab:probing_mlp}. Overall, we observe that fixed visual representations, when paired with a single linear layer, yield reasonable performance on simpler tasks such as Simple Length Comparison and Angle Comparison. However, they struggle significantly with more complex tasks like Length Comparison and Chart Projection. Therefore, we conclude that \textbf{pre-trained vision encoders do not convey sufficient information through their fixed visual representations for a linear classifier to succeed at visual arithmetic}. This may be attributed to two potential reasons: (1) the visual representations genuinely lack the information necessary for visual arithmetic tasks, or (2) a linear layer lacks the capacity to effectively leverage the visual features provided.

To further investigate the underlying cause of this limitation, we perform additional experiments by fine-tuning the LLM-based text decoder component of LLaVA-v1.5, while keeping its vision encoder frozen. In VLMs, visual representations are concatenated with text representations in the decoder. Unlike a linear layer, the text decoder can process textual queries as inputs, offering an opportunity to understand the effect of textual clues provided by input queries. We test with three different queries: an \textbf{\textsc{Original}} query reflecting the task as shown in \Cref{fig:probing_tasks}, an \textbf{\textsc{Empty}} query which is a blank string, and an \textbf{\textsc{Irrelevant}} query such as ``\textit{My name is John?}''. Additionally, we evaluate LLaVA-v1.5 in a zero-shot setting for comparisons. Given our previous observations, these experiments focus exclusively on the Length Comparison and Chart Projection tasks.
\input{tables/probing_llm}

The fine-tuned LLaVA results are shown in \Cref{tab:probing_llm}. We have the following observations. First, \textbf{existing VLMs do struggle with challenging visual arithmetic when used in zero-shot manners, achieving less than 90\% and 75\% on the two more challenging probing tasks, even with extensive in-domain training}. %
The finding is consistent with prior studies \cite{Rahmanzadehgervi_2024_blind, wang2024vdlm} and highlights the validity and complexity of our proposed probing tasks. Second, \textbf{VLMs fine-tuned on in-domain data perform reasonably well in visual arithmetic}. LLaVA-v1.5-7B is able to achieve an accuracy of above 95\% on both Length Comparison and Chart Projection tasks. Third, \textbf{the high performance of fine-tuned VLMs on in-domain data is due to the larger capacity of an LLM}. Comparing the three different queries, we see the performance on Length Comparison and Chart Projection does not vary too much. This means that a fine-tuned LLaVA-v1.5-7B performs well even when the query provides no clue or irrelevant information about the given tasks. Combing this observation with our findings in \Cref{tab:probing_mlp}, we learn that fine-tuned VLMs perform well because the LLM-based text decoder have larger capacity than a linear layer %
rather than leveraging the semantics of the input query. \looseness=-1

\subsection{Visual Representation Inspection}
\begin{figure*}[t]
    \centering
    
    \begin{subfigure}[b]{0.34\linewidth}
        \centering
        \includegraphics[width=\linewidth]{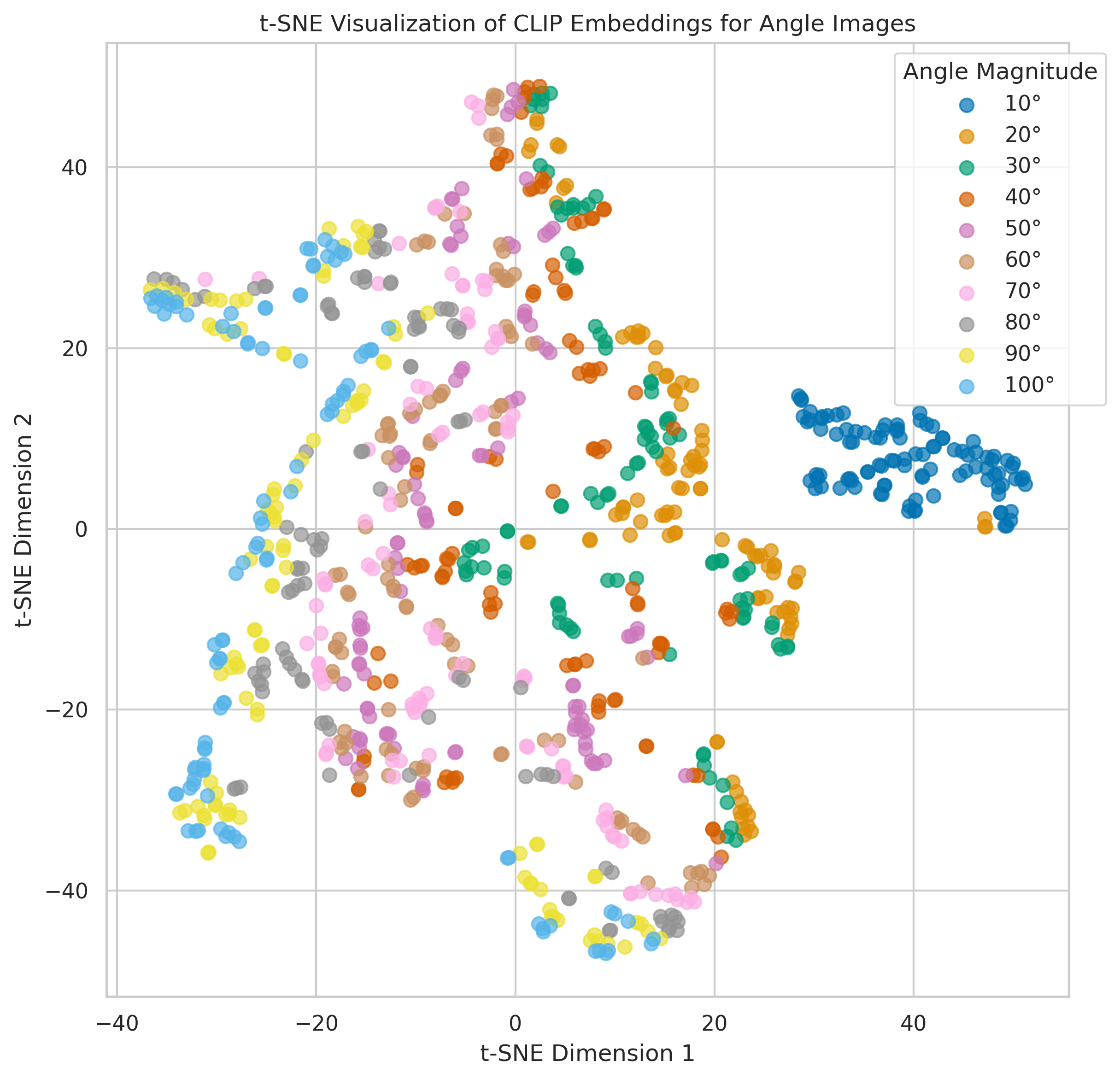}
        \caption{Angle clustering by CLIP}
        \label{fig:tsne_clip_angle_sub}
    \end{subfigure}
    \hspace{0.08\linewidth}
    \begin{subfigure}[b]{0.34\linewidth}
        \centering
        \includegraphics[width=\linewidth]{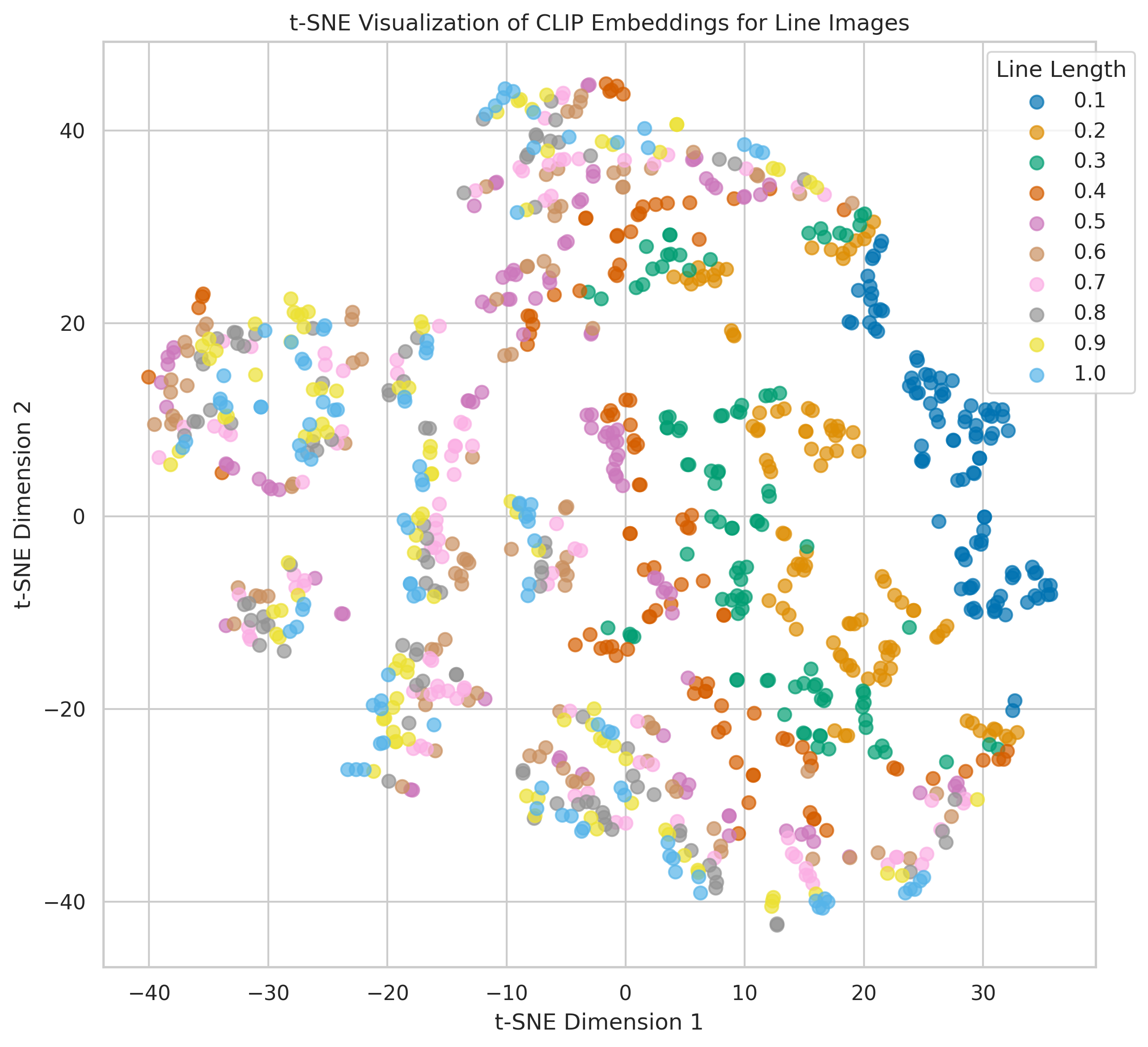}
        \caption{Line clustering by CLIP}
        \label{fig:tsne_clip_line_sub} %
    \end{subfigure}

    \begin{subfigure}[b]{0.4\linewidth}
        \centering
        \includegraphics[width=\linewidth]{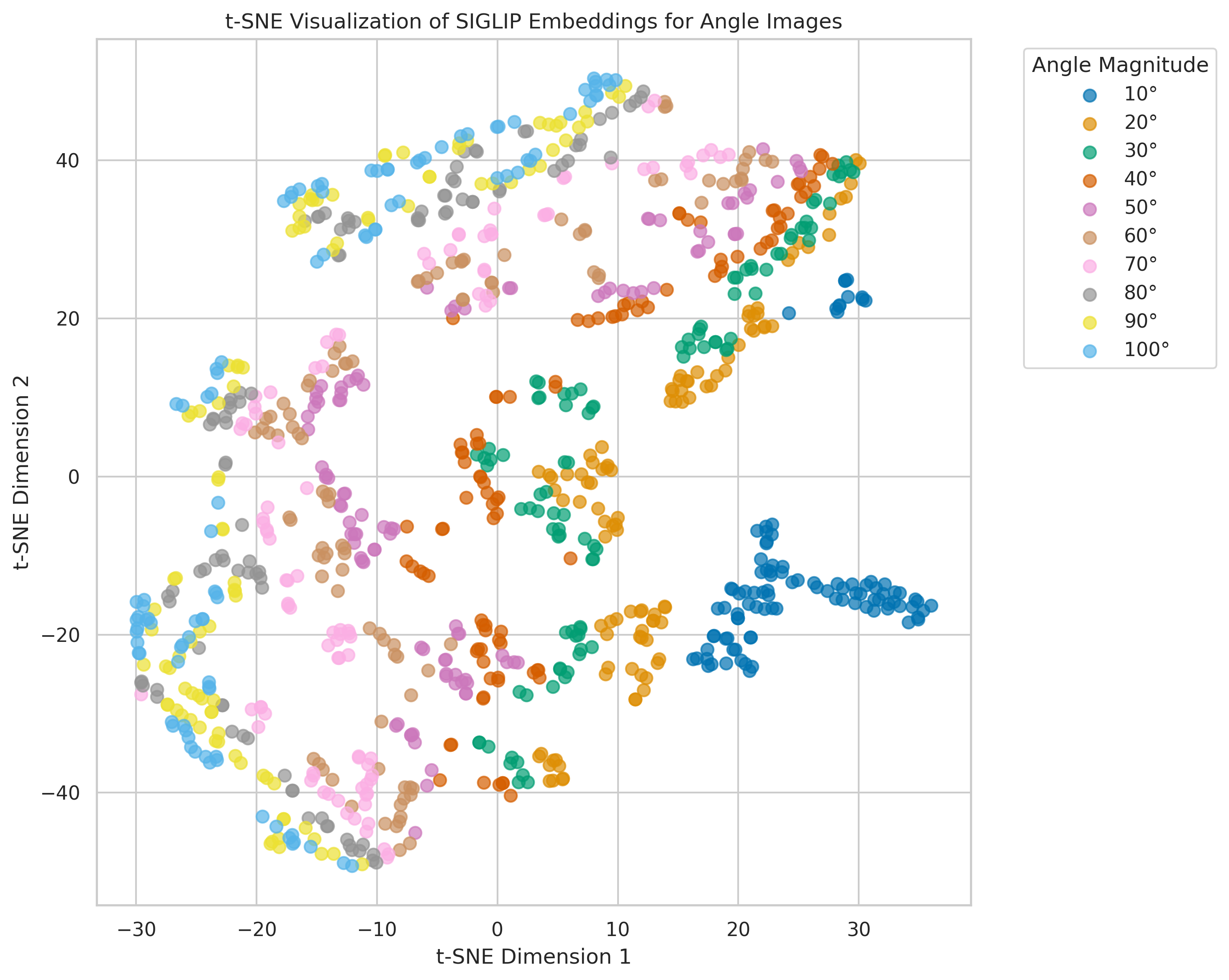}
        \caption{Angle clustering by SigLIP}
        \label{fig:tsne_siglip_angle_sub}
    \end{subfigure}
    \begin{subfigure}[b]{0.4\linewidth}
        \centering
        \includegraphics[width=\linewidth]{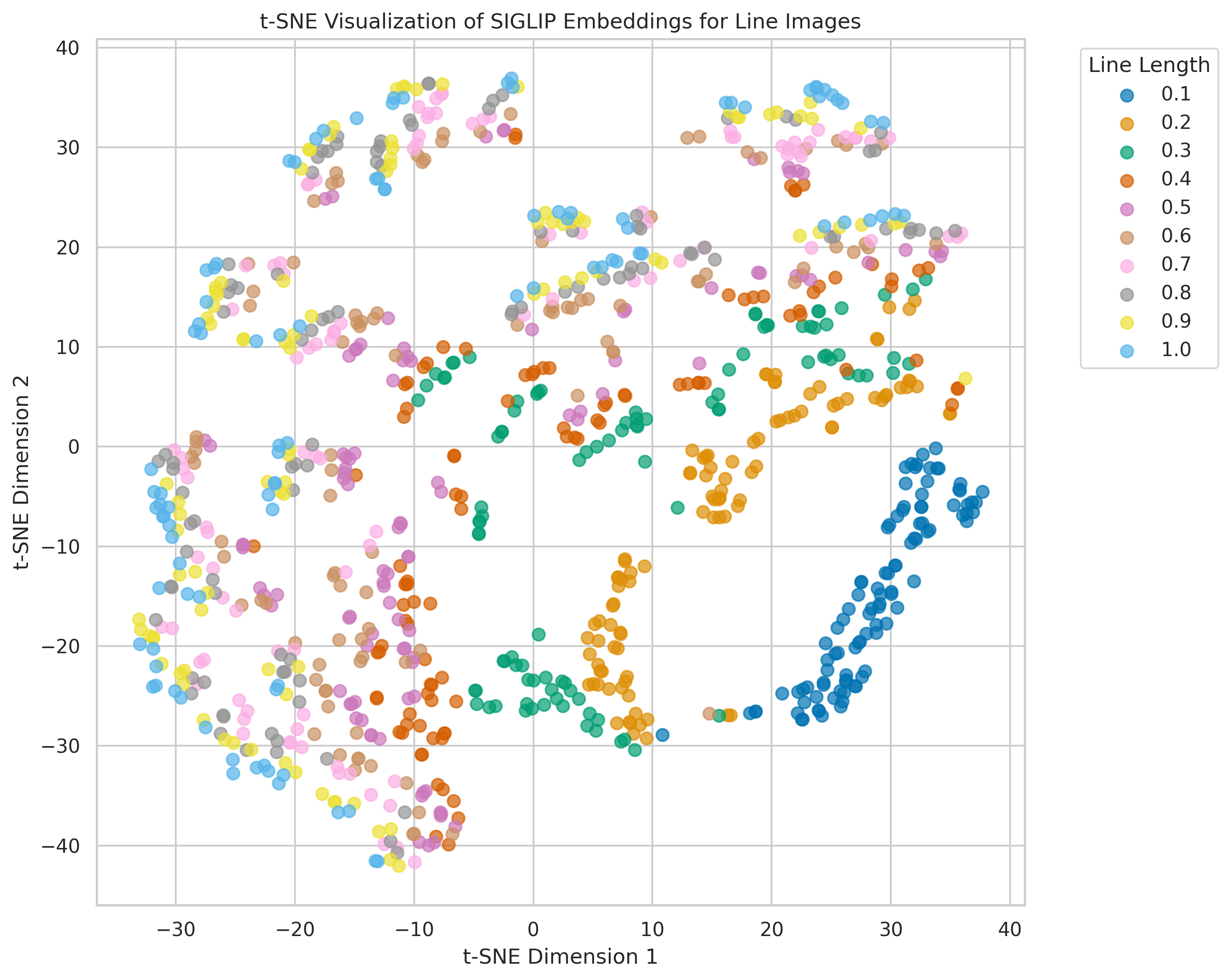}
        \caption{Line clustering by SigLIP}
        \label{fig:tsne_siglip_line_sub}
    \end{subfigure}
    \vspace{-2mm}
    \caption{t-SNE visualizations of clustering by CLIP and SigLIP for angles and lines.}
    \label{fig:tsne_combined}
    \vspace{-5mm}
\end{figure*}

While the analysis conducted in \Cref{subsec:probing_analysis} reveals important findings, one may argue that the fine-tuning success might be caused by the fine-tuned model exploiting patterns in binary tasks.
Motivated by the visual representation analysis of \citet{rudman2025forgotten}, we conduct additional experiments focusing on visual representations' ability to differentiate isolated geometric properties. We generated datasets of images, each containing a single angle (sampled in 10-degree increments from 10° to 100°) or a single line (sampled in 0.1 length increments from 0.1 to 1.0). We created 100 images per class, resulting in 1,000 images each for the angle and line datasets. We then extracted visual embeddings for these images using the pre-trained CLIP and SigLIP vision encoders. To assess whether embeddings for different angles or lengths occupy distinct regions in the feature space (a prerequisite for successful differentiation or classification), we performed t-SNE visualization \cite{vandermaaten08tsne} to understand cluster separation. The results are shown in \Cref{fig:tsne_combined}.

We have the following observations. First, there are some clear clusters for both CLIP and SigLIP embeddings, particularly for smaller angles and shorter lines. However, the separation is less distinct for longer lines and larger angles, especially with CLIP embeddings. Second, the separation for angles is more obvious than that for lines, aligning with our findings in Table 3 that Length Comparison is more challenging than Angle Comparison. Third, SigLIP embeddings demonstrate superior clustering for both angles and lines. This is also consistent with our Table 3 results where SigLIP outperforms CLIP across the proposed probing tasks. \looseness=-1

Based on these observations and our findings in \Cref{subsec:probing_analysis}, we conclude that \textbf{while visual representations from VLMs do encode information for visual arithmetic, it cannot be effectively decoded without further fine-tuning}, which may in turn affect their zero-shot performance on downstream tasks like chart understanding and that \textbf{fine-tuning success is \textit{not} merely exploiting patterns in a binary task}. \looseness=-1

%% file: tables/probing_mlp.tex
\begin{table*}[t]
    \small
    \centering
    \begin{adjustbox}{max width=0.95\textwidth}
    {
    \begin{tabular}{lcccc}
        \toprule
        
        \textbf{Vision Encoder} & \textbf{Angle Comparison} & \textbf{Perpendicular Detection} & \textbf{Length Comparison} & \textbf{Chart Projection}\\
        
        \midrule
        LLaVA-v1.5-proj & 89.7 & 87.3 & 74.4 & 61.3\\
        CLIP ViT-L/14 & 95.8 & \textbf{92.2} & 81.0 & 60.0\\
        SigLIP-SO400M/14 & \textbf{98.5} & 92.1 & \textbf{89.9} & \textbf{74.5}\\
        InternViT-300M-V2.5 & 88.2 & 85.5 & 70.8 & 59.0 \\
        DINOv2-Large & 96.2 & 94.7 & 81.8 & 57.3\\
        \midrule
        Random guessing & 50.0 & 50.0 & 50.0 & 50.0\\
        
        \bottomrule
    \end{tabular}
    }
    \end{adjustbox}
    \vspace{-2mm}
    
    \caption{Accuracy (\%) of different vision encoders with a linear classifier on the test set of each probing task. We conduct feature probing experiments by freezing the vision encoder and only fine-tuning the linear layer for binary classification. LLaVA-v1.5-proj refers to the representations obtained from the projection layer of LLaVA-v1.5.}
    \label{tab:probing_mlp}
    \vspace{-6mm}
\end{table*}

%% file: tables/probing_llm.tex
\begin{table*}[t]
    \small
    \centering
    \begin{adjustbox}{max width=0.95\textwidth}
    {
    \begin{tabular}{lcccc}
        \toprule
        
        \textbf{VLM} & \textbf{Query Type} & \textbf{Fine-tuned?} & \textbf{Length Comparison} & \textbf{Chart Projection}\\
        
        \midrule
        \multirow{5}{*}{LLaVA-v1.5-7B} & - & {\color{darkred} \xmark } & 50.0 & 51.3\\
        \cmidrule{2-5}
         & \textsc{Original} & {\color{lightgreen} \cmark} & 95.4 & \textbf{98.9}\\
         & \textsc{Empty} & {\color{lightgreen} \cmark} & 95.2 & 98.1\\
         & \textsc{Irrelevant} & {\color{lightgreen} \cmark} & \textbf{95.8} & 97.8\\

        \bottomrule
    \end{tabular}
    }
    \end{adjustbox}
    \vspace{-2mm}
    
    \caption{Accuracy (\%) of fine-tuned LLaVA-v1.5-7B with different queries on the test set of each probing task. We conduct experiments by freezing the vision encoder and only fine-tune the LLM decoder for binary classification. The top row displays the zero-shot performance.}
    \label{tab:probing_llm}
    \vspace{-6mm}
\end{table*}

%% file: content/04_method.tex
\section{\method~}
\label{sec:method}

\begin{figure*}[t]
    \centering
    \includegraphics[width=1.0\linewidth, trim=20 50 20 40, clip]{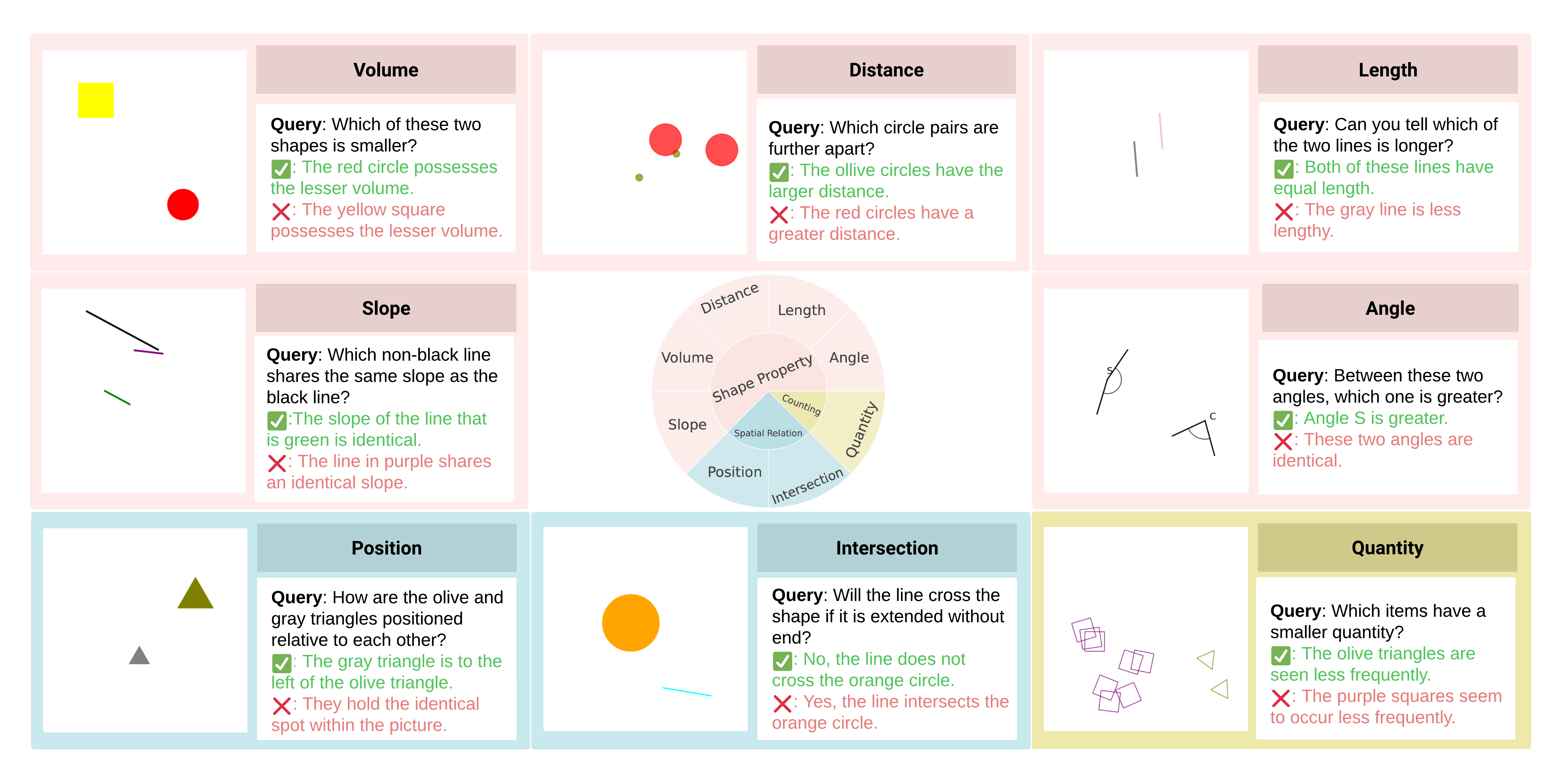}
    \vspace{-7mm}
    \caption{Example training data for \method~. Each example consists of a visual input, a query prompting comparison of a specific property (i.e. angle, length, distance, and etc), a positive response consistent with the visual input, and a negative response that contradicts it.}
    \vspace{-5mm}
    \label{fig:training_data}
\end{figure*}

To address the challenges VLMs face in performing visual arithmetic, we propose a novel post-training method inspired by Piaget's theory of cognitive development \cite{piaget1952origins}, which outlines four stages: Sensorimotor, Preoperational, Concrete Operational, and Formal Operational. Each stage represents a different ability to process information and solve problems, culminating in abstract reasoning. The Concrete Operational Stage is particularly relevant. At this stage, children develop (1) \textit{conservation}, understanding that certain properties like length remains constant despite changes in appearance, and (2) \textit{decentration}, the ability to consider multiple aspects of a situation at once. These skills are essential for VLMs to perform visual arithmetic accurately, recognizing invariant properties such as length or angle across transformations. Current VLM training paradigms often neglect these cognitive processes, resulting in models that struggle to maintain key properties during visual transformations and to integrate multiple visual features effectively. While pre-trained visual encoders use losses that encourage some invariance to transformations, the integration of vision and language representations in decoders often lacks explicit enforcement of conservation and decentration principles, leading to models that capture visual features but fail to reason about them effectively.

To address these issues, we present a post-training method, Cognitive Alignment (\method~), aimed at enhancing VLMs' understanding of \textit{conservation} and \textit{decentration}. Our approach %
explicitly trains VLMs to recognize invariant properties like length, angle, and count across different visual transformations. We achieve this by presenting the model with pairs of figures and associated queries designed to highlight these properties. The queries prompt the model to compare and contrast the figures, focusing on whether a specific property is different or same despite variations in appearance. This approach encourages the model to develop a stronger understanding of geometric concepts and move beyond superficial visual comparisons.
Furthermore, we leverage DPO \cite{rafailov2023dpo} for training, rather than SFT. DPO allows the model to learn from both positive and negative examples within the preference framework, providing a richer learning signal compared to SFT. By strengthening these cognitive capacities within VLMs, our goal is to improve their performance on tasks involving visual arithmetic. The subsequent subsections detail the specific training procedure employed (\Cref{subsec:training_setting}) and the automated construction of our training data (\Cref{subsec:training_collect}).

\subsection{DPO Training Objective}

\label{subsec:training_setting}
Our goal is to train a model with parameters $\theta$ that learns \textit{conservation} and \textit{decentration} from contrasting responses by maximizing the conditional probability of positive responses over their negative counterparts. Concretely, the DPO training data consists of preference pairs, each containing a user query $Q$, an input image $I$, a positive response $R_p$ and a negative response $R_n$. The entire set of DPO training data can be represented as $\mathcal{D}=\left\{ \left(Q,I,R_p,R_n \right)^{(i)} \right\}_{i=1}^{\lvert \mathcal{D} \rvert}$. The objective function $\mathcal{L}_\mathrm{DPO}$ that DPO minimizes is:

\begin{align*}
\label{eq:dpo}
\small
&\mathcal{L}_\mathrm{DPO}(\pi_\theta;\pi_\mathrm{ref}) =-\mathrm{E}_{(Q,I, R_p,R_n)\sim \mathcal{D}}\Big[\log \sigma (r_{\Delta})\Big],\\
&r_{\Delta} = \beta \log\frac{\pi_\theta(R_p\vert Q, I)}{\pi_\mathrm{ref}(R_p\vert Q, I)} - \beta\log\frac{\pi_\theta(R_n\vert Q, I)}{\pi_\mathrm{ref}(R_n\vert Q, I)},
\end{align*}
where $\sigma$ is the sigmoid function%
, $\pi_{\theta}$ is the parameterized policy under training%
, $\pi_\mathrm{ref}$ is the initial frozen %
policy, and $\beta$ is a hyper-parameter that controls the deviation from $\pi_\mathrm{ref}$.

\subsection{Training Data Synthesis}
\label{subsec:training_collect}

To effectively train VLMs on the principles of \textit{conservation} and \textit{decentration}, we require a training dataset designed to highlight these concepts. This section details our automated process for synthesizing training data, encompassing visual generation and tailored query-response construction. We draw inspiration from our probing tasks but adapt the format to better suit the DPO training procedure. By plotting two shapes within a single image, we allow the model to directly compare invariant properties like length and angle across various transformations. We devise eight fundamental tasks, each of which aim to enhance VLMs' different abilities to reason about visual arithmetic operations: understanding \textit{angle}, \textit{length}, \textit{distance}, \textit{quantity}, \textit{volume}, \textit{position}, \textit{slope}, and \textit{intersection}. An overview of these tasks is shown in \Cref{fig:training_data}

\input{tables/probing_ours}

To automate data synthesis, we present a data generation pipeline. First, we programmatically generate images using Python, allowing precise control over each figure's properties, such as lengths and positions. Next, we create query-response pairs for these images based on predefined templates (see \Cref{tab:query_templates}). These queries are designed to prompt VLMs to make comparisons, identify similarities/dissimilarities, and reason about geometric properties. Given the known ground truth, positive responses are generated by accurately populating placeholders in the templates, while negative responses are created with incorrect values. For instance, a positive query for the first sub-figure in \Cref{fig:training_data} might be, “\textit{The angle S is larger.}”, and a negative query might be, “\textit{The angle C is larger.}” To ensure diversity, we use an LLM\footnote{\texttt{gpt-4o} is used for paraphrasing.} to create multiple variations of each query and response, following the approach of \citet{huang-etal-2024-crmarena}. We synthesize a total of 64,000 training instances for DPO, with a balanced splits of each task.

\subsection{Effectiveness on the Probing Tasks}

To assess the effectiveness of \method~ on our proposed probing tasks, we trained three VLMs with varying scales and architectures: LLaVA-OV-0.5B \cite{li2024llavaov}, InternVL-2.5-MPO-1B \cite{wang2024mpo}, and InternVL-2.5-MPO-4B using \method~, as described in \Cref{subsec:training_setting} and \Cref{subsec:training_collect}, for one epoch. The results are presented in Table 3.%

We observe that \method~ demonstrably improves performance across all three models across all probing tasks. More significant gains are observed on simpler tasks of Angle Comparison, likely due to their similarity with the DPO training instances. For instance, LLaVA-OV-0.5B sees a substantial 54.0\% improvement on Angle Comparison after training with \method~.  This highlights the effectiveness of our approach in enhancing the core visual arithmetic capabilities that are crucial for these tasks. Interestingly, while the gain in angle-related tasks like Angle Comparison was substantial, the performance increases for Perpendicularity Detection were more modest (e.g., a 2.0\% improvement for LLaVA-OV-0.5B). This suggests that certain geometric properties, such as perpendicularity, may pose greater challenges. %

Overall, these findings collectively demonstrate the effectiveness of \method~ in enhancing visual arithmetic capabilities across various VLMs.%

%% file: tables/probing_ours.tex
\begin{table*}[t]
    \small
    \centering
    \begin{adjustbox}{max width=0.95\textwidth}
    {
    \begin{tabular}{lcccc}
        \toprule
        
        \textbf{Model} & \textbf{Angle Comparison} & \textbf{Perpendicular Detection} & \textbf{Length Comparison} & \textbf{Chart Projection}\\
        
        \midrule
        
        LLaVA-OV-0.5B & 51.8 & 50.5 & 52.5 & 50.7\\
        ~~~ +\method~ (Ours) & \textbf{79.8} \markgreen{(+54.0\%)} & \textbf{51.5} \markgreen{(+2.0\%)} & \textbf{53.4} \markgreen{(+1.7\%)} & \textbf{53.2} \markgreen{(+4.9\%)}\\
        \hdashline\noalign{\vskip 0.5ex}
        InternVL-2.5-MPO-1B & 51.8 & 49.6 & 52.3 & 60.0\\
        ~~~ +\method~ (Ours) & \textbf{52.1} \markgreen{(+0.6\%)} & \textbf{50.7} \markgreen{(+2.0\%)} & \textbf{52.6} \markgreen{(+0.6\%)} & \textbf{66.9} \markgreen{(+11.5\%)}\\
        
        \hdashline\noalign{\vskip 0.5ex}
        InternVL-2.5-MPO-4B & 60.6 & 54.9 & 56.3 & 84.0\\
        ~~~ +\method~ (Ours) & \textbf{72.3} \markgreen{(+19.3\%)} & \textbf{56.4} \markgreen{(+2.7\%)} & \textbf{60.0} \markgreen{(+6.6\%)} & \textbf{86.3} \markgreen{(+2.7\%)}\\
        
        \bottomrule
    \end{tabular}
    }
    \end{adjustbox}
    \vspace{-2mm}
    
    \caption{Accuracy (\%) of different VLMs on our proposed probing tasks. All models produce output in a zero-shot fashion without fine-tuning on the tasks.}%
    \label{tab:probing_ours}
    \vspace{-5mm}
\end{table*}

%% file: content/05_results.tex
\section{Generalizability of \method~}

\label{sec:exps}
Now that we have demonstrated the advantage of \method~ on our probing tasks, we ask: \textbf{\textit{does the improvement on simple visual arithmetic tasks transfer to more complex tasks?}} To answer this question, we explore whether \method~ enhances model performance in chart understanding and geometric problem-solving. In the following subsections, we detail the experimental setup (\Cref{subsec:exp_setup}) and present our findings (\Cref{subsec:results}).

\subsection{Experimental Setups}
\label{subsec:exp_setup}

\paragraph{Benchmarks}
We evaluate the effectiveness of our method on two tasks relevant to visual arithmetic: chart understanding and geometry problem-solving. For chart understanding, we utilize the \chocolate~ dataset \cite{huang-etal-2024-lvlms}, which tests a model's capability to determine whether a given caption is factually consistent with its corresponding chart.\footnote{We decided against using other common datasets like ChartQA \cite{masry-etal-2022-chartqa} due to their training data already being included in some VLMs, such as LLaVA-OneVision.} \chocolate~ comprises three splits: \textsc{Lvlm}, \textsc{Llm}, and \textsc{Ft}, each generated by models of varying architectures and scales. Each \chocolate~ instance is annotated with a binary label $\mathcal{L} \in \{\texttt{consistent}, \texttt{inconsistent}\}$. The dataset includes a total of 1,187 chart-caption pairs. For geometry problem-solving, we assess performance using the test set of the \mathv~ dataset \cite{wang2024mathv}, which comprises 3,040 questions spanning 16 mathematical disciplines. We concentrate on the eight disciplines related to geometry: analytic geometry (\textsc{AnaG}), combinatorial geometry (\textsc{CombG}), descriptive geometry (\textsc{DescG}), solid geometry (\textsc{SolG}), transformation geometry (\textsc{TransG}), and three metric geometry branches - angle, area, and length. For evaluations, we employ AUC score for \chocolate~ and accuracy for \mathv~, in alignment with \citet{huang-etal-2024-lvlms} and \citet{wang2024mathv}. Detailed dataset statistics for these benchmarks are provided in \Cref{apx:dataset_stats}.

\input{tables/main}

\paragraph{Models and Baselines}

To assess the efficacy of \method~ compared to methods that directly optimize model capabilities towards specific tasks, we consider a chart supervised fine-tuning dataset: \textsc{ChartGemma160k} \cite{masry2024chartgemma}, as well as one geometric problem-solving dataset: \textsc{Geo170K} \cite{gao2023gllava}. 
We use the above methods to train three open-source VLMs for one epoch: InternVL2.5-1B-MPO \cite{wang2024mpo}, InternVL2.5-4B-MPO \cite{wang2024mpo}, and LLaVA-OV-0.5B \cite{li2024llavaov}. %
We also compare performance of two VLMs instruction-tuned specifically for chart understanding and geometric problem-solving: ChartGemma-3B \cite{masry2024chartgemma} and G-LLaVA-13B \cite{gao2023gllava}. Experimental details can be found in \Cref{apx:training_setting}.

\subsection{Results}
\label{subsec:results}

The results for experiments on \chocolate~ and \mathv~ are shown in \Cref{tab:main_results}. We find that \textbf{\method~ is effective in enhancing chart understanding and geometric problem-solving capabilities of VLMs even though \method~ was not specifically optimized for these two tasks.} On average, \method~ boosts the performance by 4.6\% and 2.9\% on the \chocolate~ and \mathv~ datasets, respectively. This shows that patching fundamental capabilities such as visual arithmetic of VLMs can enhance their capabilities in tasks involving such abilities. %

More importantly, we find that \textbf{\method~ demonstrates better generalizability compared to supervised fine-tuning VLMs using task-specific data}. For instance, when comparing the InternVL-2.5-MPO-1B variants, \method~ achieves an average score of 61.5\% on \chocolate~, outperforming both the \textsc{ChartGemma160K} (59.1\%) and \textsc{Geo170K} (59.2.\%) variants. Similarly, on the \mathv~ dataset, while the \textsc{Geo170K} variant shows competitive performance, \method~ achieves a comparable average performance across all geometry subtasks, indicating a broader improvement.  Notably, \method~ requires only 60\% less training data compared to these two baseline methods. %

The results suggest that \method~ offers a valuable approach to enhancing VLMs by improving their fundamental visual arithmetic capabilities. It exhibits strong generalizability across different tasks and base models, often outperforming or achieving comparable performance to task-specific fine-tuning methods without being explicitly trained on the target datasets. This highlights the potential of focusing on foundational skills to unlock broader capabilities in VLMs.

\subsection{Discussions}

\begin{figure}[t]
    \centering
    \includegraphics[width=0.98\linewidth, trim=0 0 0 5, clip]{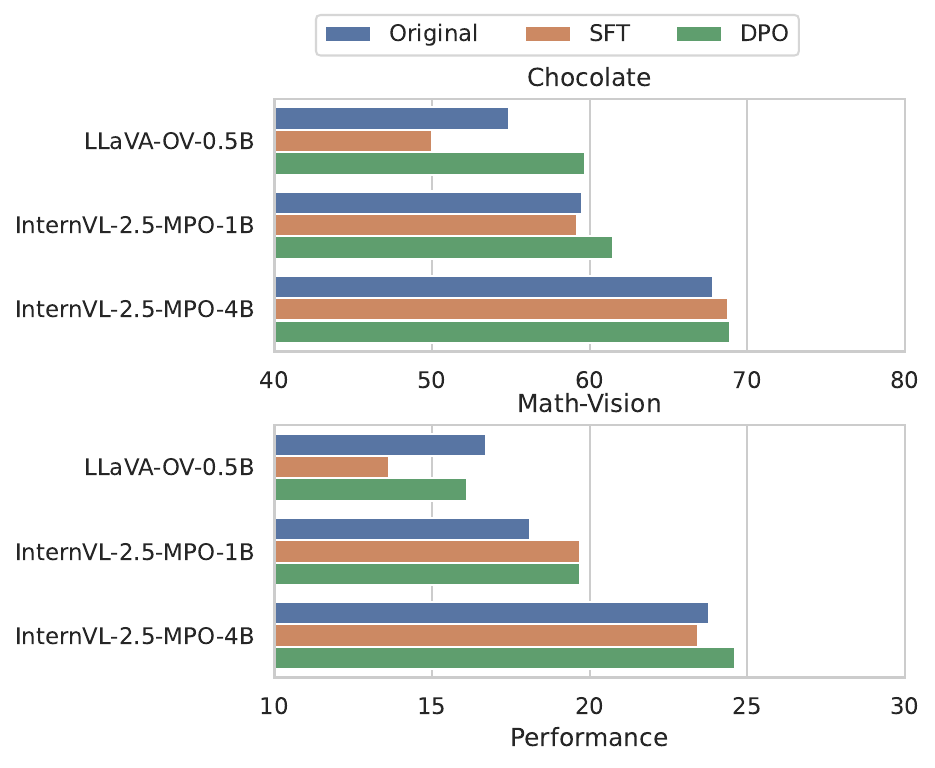}
    \vspace{-4mm}
    \caption{Performance comparison when training different models with SFT and DPO.}
    \vspace{-5mm}
    \label{fig:sft_vs_dpo}
\end{figure}

\paragraph{Impact of learning from contrasting examples}
We investigate the impact of learning from contrasting examples versus solely positive examples by comparing DPO (the default \method~ setting) and SFT training method (using only the positive response). \Cref{fig:sft_vs_dpo} presents the results. We observe that the SFT approaches can lead to much worse performance (e.g. LLaVA-OV-0.5B), while the DPO approach improves performance over the original models more consistently. This suggests that learning from contrasting examples provides a richer learning signal compared to traditional supervised learning, leading to better performance. \looseness=-1

\paragraph{Impact on general VLM benchmarks}

To assess the impact of \method~ on general VLM capabilities, we compare the performance on two additional benchmarks: MME \cite{fu2023mme} and MMMU \cite{yue2024cvpr}. The results are presented in \Cref{fig:general_benchamrk}. Overall, \method~ consistently improves performance across most settings (five out of six), indicating that its benefits extend beyond the specific probing tasks and generalize to other multimodal reasoning challenges. This suggests that \textbf{\method~ enhances visual arithmetic capabilities without compromising performance on general tasks.}

%% file: tables/main.tex
\begin{table*}[t]
    \small
    \centering
    \begin{adjustbox}{max width=0.95\textwidth}
    {
    \begin{tabular}{lccccccccccccc}
        \toprule
        
          & \multicolumn{4}{c}{\textbf{\chocolate~}} & \multicolumn{9}{c}{\textbf{\mathv~}}\\
        \cmidrule(lr){2-5} \cmidrule(lr){6-14}
        \textbf{Model}   & \textsc{Lvlm} & \textsc{Llm} & \textsc{Ft} & AVG & \textsc{AnaG} &\textsc{CombG} & \textsc{DescG} & \textsc{Angle} & \textsc{Area} &  \textsc{Len} & \textsc{SolG} & \textsc{TransG} & AVG\\
        \midrule
        GPT-4o & 74.7 & 69.6 & 68.5 & 70.9 & 35.7 & 31.2 & 28.8 & 38.2 & 34.8 & 36.7 & 31.1 & 21.4 & 32.2 \\
        LLaVA-Next-Vicuna-7B & 50.0 & 50.3 & 50.1 & 50.1 & 9.5 & 14.9 & 18.3 & 12.7 & 15.6 & 13.4 & 7.0 & 17.3 & 15.6\\
        Qwen2-VL-7B-Instruct & 57.3 & 64.0 & 71.1 & 65.4 & 17.9 & 17.9 & 24.0 & 19.7 & 22.4 & 21.4 & 16.0 & 25.0 & 19.5 \\
        ChartGemma-3B & 51.8 & 54.2 & 53.7 & 53.2 & 11.9 & 13.6 & 14.4 & 9.8& 11.2 & 10.0 & 10.6 & 13.7 & 11.9\\
        G-LLaVA-13B & 50.0 & 50.0 &	50.0 & 50.0  & 14.3 & 15.9 & 22.1 & 19.1 & 20.0 & 21.2 & 15.6 & 16.1 & 18.0\\
        \midrule
         LLaVA-OV-0.5B & 56.6 & 50.4 & \textbf{57.8}   & 54.9 & \textbf{16.7} & 17.2 & \textbf{22.1} & 17.3 & 13.6 & \textbf{18.9} & 11.1 & 16.7 & \textbf{16.7}\\
         ~~~ +\textsc{ChartGemma160k} & 50.2 & 50.0 &  50.3 & 50.2 & 11.9 & 15.6 & 16.3 & \textbf{17.9} & 12.4 & 14.6 & 10.2 & \textbf{19.0} & 14.7\\
         ~~~ +\textsc{Geo170K}  & 50.8 & 48.7 & 50.5 & 50.0 & 16.7 & 16.9 & 13.5 & 17.3 & \textbf{14.6} & 16.9 & 10.7 & 17.3 & 15.5 \\
         ~~~ +\method~ (Ours) & \textbf{56.7} & \textbf{64.7} & 57.7 & \textbf{59.7}  & 15.5 & \textbf{17.5} & 19.2 & 17.3 & 13.8 & 17.8 & \textbf{11.5} & 16.1 & 16.1\\

         \hdashline\noalign{\vskip 0.5ex}
         
         InternVL-2.5-MPO-1B & 53.2 & 60.2 & 65.0 & 59.5 & 16.7 & 18.2 & 22.1 & 26.0 & 15.8 & 18.0 & 10.7 & 17.3 & 18.1\\
         ~~~ +\textsc{ChartGemma160k} & 54.5 & 61.8 & 60.9 & 59.1 & 20.2 & 16.6 & 24.0 & \textbf{26.6} & 19.2 & 16.3 & 12.3 & \textbf{21.4} & 19.6\\
         ~~~ +\textsc{Geo170K} & 54.6 & 62.1 & 60.9 & 59.2 & 19.0 & 16.6 & \textbf{32.7} & 24.9 & \textbf{20.0} & 15.4 & 13.1 & 18.5 & \textbf{20.0} \\
         ~~~ +\method~ (Ours) & \textbf{59.7} & \textbf{60.1} & \textbf{64.6} & \textbf{61.5}  & 16.7 & 16.2 & 31.7 & 25.4 & 17.0 & 17.4 & \textbf{13.1} & 20.8 & 19.7 \\
        \hdashline\noalign{\vskip 0.5ex}

         InternVL-2.5-MPO-4B & 60.3 & 67.2 & 75.9 & 67.8 &  28.6 & \textbf{23.1} & 22.1 & 32.4 & 22.6 &24.9 & \textbf{19.7} & 17.9 & 23.8\\
         ~~~ +\textsc{ChartGemma160k} & \textbf{62.1} & 66.0 & 76.2 & 68.1 & 23.8 & 18.5 & 18.3 & 16.8 & 24.4 & 23.4 & 10.7 & 17.9 & 22.3 \\
         ~~~ +\textsc{Geo170K} & 60.0 & 65.5 & 64.3 & 59.9 & 32.1 & 18.8  & 23.7 & 31.2 & \textbf{23.6} & 23.8 & 15.6 & 21.4 & 23.8\\
         ~~~ +\method~ (Ours) & 61.2 & \textbf{68.6} & \textbf{76.8}  & \textbf{68.9} & 27.4 & 19.8 & \textbf{24.0} & \textbf{32.9} & 22.6 & \textbf{26.9} & 17.2 & \textbf{25.6} & \textbf{24.6}\\

        \bottomrule
    \end{tabular}
    }
    \end{adjustbox}
    \vspace{-2mm}
    
    \caption{Performance (\%) on the \chocolate~ and \mathv~ datasets. }
    \label{tab:main_results}
    \vspace{-6mm}
\end{table*}

%% file: content/02_related_work.tex
\section{Related Works}

\subsection{Vision Language Models}
Vision language models (VLMs) are multimodal models that learns to generate text outputs based on both visual and textual inputs. The development of large-scale VLMs has demonstrated impressive zero-shot capabilities, enabling them to perform well with a variety of image types, such as documents and web pages \cite{liu2023llava, dai2023instructblip, Achiam2023GPT4TR, gemini, claude}. These VLMs generally consist of three major components: a vision encoder, such as CLIP \cite{Radford2021LearningTV} or SigLIP \cite{Zhai_2023_siglip}, which processes visual inputs; a language model that handles textual inputs and generates text tokens; and a projector layer that connects the image and text modalities. Typically, VLMs are trained using image captioning data and instruction-tuning datasets. Recently, several post-training strategies have been suggested to enhance VLM capabilities in areas like conversational interaction \cite{xiong2024llavaovchat} and reasoning \cite{wang2024mpo}. In this work, we propose a new post-training strategy, \method~, for improving VLMs' proficiency in understanding visual arithmetic operations. \looseness=-1

\subsection{Shortcomings of Vision Language Models}

While Vision-Language Models (VLMs) demonstrate impressive performance across a range of tasks, several studies have highlighted their limitations by examining various aspects such as architectures \cite{McKinzie2024MM1, Karamcheti2024Prismatic, Tong_2024_eyes, shi2025when}, training methods \cite{laurencon2024what}, and data considerations \cite{udandarao2024no, gadre2024datacomp,zhang2024why,wei2024slow}. Some research indicates that VLMs struggle with specific tasks, including basic geometric understanding \cite{gao2023gllava,ullman2024illusion} and chart comprehension \cite{huang-etal-2024-lvlms}, and are prone to hallucinations \cite{qiu2024valor}.  Our study seeks to uncover the root causes behind these challenges, especially those that involve visual arithmetic operations, and proposes solutions to address these shortcomings.  \looseness=-1

%% file: content/06_conclusion.tex
\begin{figure}[t]
    \centering
    \includegraphics[width=0.98\linewidth, trim=0 0 0 5, clip]{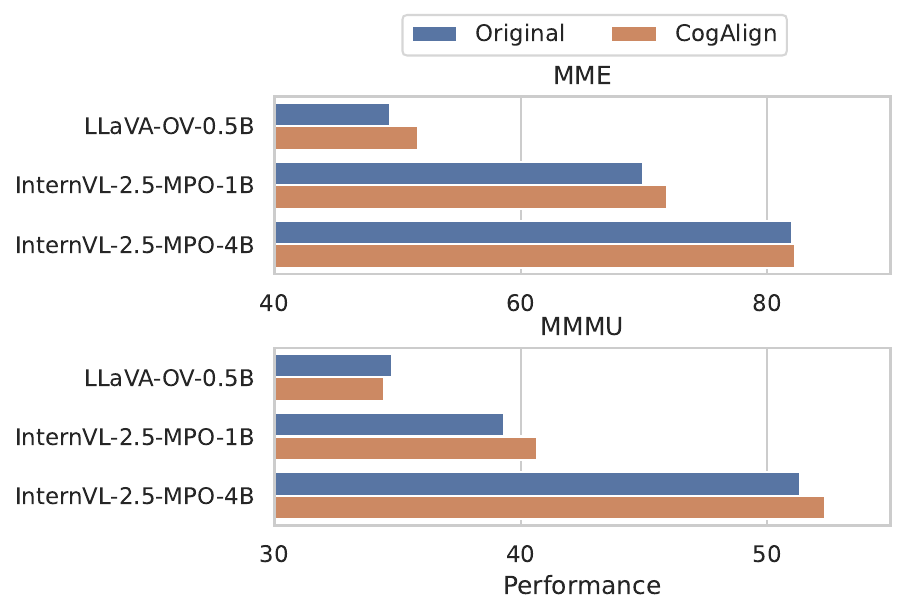}
    \vspace{-4mm}
    \caption{Performance on two general VLM benchmarks: MME and MMMU with or without \method~.}
    \vspace{-5mm}
    \label{fig:general_benchamrk}
\end{figure}
\section{Conclusion}

This study investigates the challenges faced by VLMs in performing visual arithmetic, revealing that while visual encoders often capture necessary information, text decoders struggle to effectively utilize it.  We introduce \method~, a novel post-training strategy inspired by Piaget's theory of cognitive development, focusing on enhancing VLMs' understanding of conservation and decentration through DPO training. Our evaluations show that \method~ not only enhances VLMs' understanding of visual arithmetic, but also improves their performance in chart understanding and geometric problem-solving through experiments on the \chocolate~ and \mathv~ datasets, showcasing its effectiveness and generalizability across various models and tasks. Notably, \method~ often outperforms or achieves comparable results to task-specific supervised fine-tuning methods without direct training on the target domain, highlighting the potential of bolstering foundational cognitive skills for broader VLM capabilities. Future work could explore how \method~ impacts other multimodal tasks beyond charts and geometry, potentially leading to a more unified approach in VLM training where generalizability is prioritized. \looseness=-1

%% file: content/07_limitation.tex
\section{Limitations}

\paragraph{Probing Tasks} While the probing tasks we have proposed provide valuable insights into the visual arithmetic capabilities of VLMs, it is important to acknowledge that they may not encompass all possible dimensions of visual reasoning. Our choice to limit the scope of these tasks was intentional, as they serve as initial, simple tests to determine whether VLMs exhibit failure in fundamental aspects of visual arithmetic. These tasks allow us to iterate different experiments in a controlled and efficient manner, providing clear, actionable insights without the complexity that more comprehensive tasks might introduce. However, there is potential to explore additional tasks that involve more complex interactions of basic geometric properties. For instance, tasks requiring the model to simultaneously assess both length and angle, or combinations of length and area, could be valuable for understanding the compositionality of these atomic tasks. \looseness=-1

\paragraph{Training Data Synthesis}The training data synthesis method of \method~ is not only scalable but also effectively enhances the visual arithmetic capabilities of VLMs. Our approach serves as a proof-of-concept, demonstrating the potential of automated data generation for improving models' understanding of basic geometric properties. To further enrich the training data, we could consider utilizing additional configurations for each task. For instance, in generating positive and negative responses, we could leverage LLMs to produce rationales based on the specific configuration of each figure. By including explanations or justifications for why a particular geometric property holds or does not hold, we could foster deeper understanding within the VLMs. \looseness=-1

%% file: content/appendix.tex
\clearpage
\appendix

\section{Dataset statistics}
\label{apx:dataset_stats}
\Cref{tab:dataset_stats_chocolate} and \Cref{tab:dataset_stats_mathv} show the dataset statistics for the \chocolate~ and \mathv~ datasets, respectively.

\input{tables/dataset_stats_chocolate}
\input{tables/dataset_stats_mathv}

\input{tables/query_templates}

\section{Training Settings}

\label{apx:training_setting}
For all models and all training approaches, we set all other hyper-parameters according to the guidelines described in their corresponding GitHub repository. These settings are described in \Cref{tab:training_setting}.

\input{tables/training_settings}

%% file: tables/dataset_stats_chocolate.tex
\begin{table}[b]
    \small
    \centering
    \begin{adjustbox}{max width=0.47\textwidth}
    {
    \begin{tabular}{lccc}
        \toprule
        
        & \textbf{\# Factual} & \textbf{\# Non-factual} & \textbf{\# Total}\\
        \midrule
        Sentence & 2,561 & 2,762 & 5,323\\
        Caption & 213 & 974 & 1,187 \\

        \bottomrule
    \end{tabular}
    }
    \end{adjustbox}
    \caption{Statistics of the \chocolate~ dataset. A sentence is considered factual if and only if it does not contain any factual error. A caption is considered factual if all its sentences are factual.\looseness=-1} 
    \label{tab:dataset_stats_chocolate}
    
\end{table}

%% file: tables/dataset_stats_mathv.tex
\begin{table}[t]
    \small
    \centering
    \begin{adjustbox}{max width=0.47\textwidth}
    {
    \begin{tabular}{lccc}
        \toprule
        
       \textbf{Statistic} & \textbf{Number} \\
         \midrule
          Total questions & 3,040 \\
          ~- multiple-choice questions & 1,532 (50.4\%) \\
          ~- Free-form questions & 1,508 (49.6\%) \\
          ~- Questions in the testmini set & 304 (10.0\%) \\
        \bottomrule
    \end{tabular}
    }
    \end{adjustbox}
    \caption{Statistics of the \mathv~ dataset. \looseness=-1} 
    \label{tab:dataset_stats_mathv}
    
\end{table}

%% file: tables/query_templates.tex
\begin{table*}[b!]
\vspace{-3mm}
\centering
 \small
 
 \resizebox{0.9\linewidth}{!}{
    \begin{tabular}{ll}
        \toprule
        \textbf{Task}  & \textbf{Query Templates}\\
        \midrule
        \multirow{5}{*}{Angle} & \texttt{1. The angle with the [COLOR] color is larger.} \\
        & \texttt{2. The angle X is larger.}\\
        & \texttt{3. The angle with the [COLOR] color is smaller.} \\
        & \texttt{4. The angle X is smaller.}\\
        & \texttt{5. These two angles are the same.}\\
        \midrule
        \multirow{5}{*}{Length} & \texttt{1. The line with the [COLOR] color is longer.} \\
        & \texttt{2. The line X is longer.}\\
        & \texttt{3. The line with the [COLOR] color is shorter.} \\
        & \texttt{4. The line X is shorter.}\\
        & \texttt{5. These two lines are the same length.}\\
        \midrule
        \multirow{3}{*}{Distance} & \texttt{1. The pair of circles with the [COLOR] color has the longer distance.} \\
        & \texttt{2. The pair of circles with the [COLOR] color has the smaller distance.} \\
        & \texttt{3. These two pair of circles have the same distance.}\\
        \midrule
        \multirow{3}{*}{Quantity} & \texttt{1. The [COLOR] [SHAPE] appears more times.} \\
        & \texttt{2. The [COLOR] [SHAPE] appears less times.} \\
        & \texttt{3. The [COLOR-A] [SHAPE-A] and [COLOR-B] [SHAPE-B] appear the same number of times.}\\
        \midrule
        \multirow{3}{*}{Volume} & \texttt{1. The [COLOR] [SHAPE] has the larger volume.} \\
        & \texttt{2. The [COLOR] [SHAPE] has the smaller volume.} \\
        & \texttt{3. These two shapes have the same volume.}\\
        \midrule
        \multirow{3}{*}{Slope} & \texttt{1. The line with the [COLOR] has the same slope.} \\
        & \texttt{2. Both lines have the same slope as the black line.} \\
        & \texttt{3. Neither line has the same slope as the black line.}\\
        \midrule
        \multirow{3}{*}{Position} & \texttt{1. The [COLOR-A] [SHAPE-A] is [POSITION] of [COLOR-B] [SHAPE-B].} \\
        & \texttt{2. They occupy the exact same position in the image.} \\
        & \texttt{3. The [COLOR-A] [SHAPE-A] is [WRONG-POSITION] of [COLOR-B] [SHAPE-B].}\\
        \midrule
        \multirow{2}{*}{Intersection} & \texttt{1. Yes, the line does intersect the [COLOR] [SHAPE].} \\
        & \texttt{2. No, the line does not intersect the [COLOR] [SHAPE].} \\

        \bottomrule
    \end{tabular}
    }
    \vspace{-2mm}
    \caption{The full set of query templates used for query generation.}
\vspace{-3mm}
\label{tab:query_templates}
\end{table*}

%% file: tables/training_settings.tex
\begin{table*}[t]
    \small
    \centering
    \begin{adjustbox}{max width=0.95\textwidth}
    {
    \begin{tabular}{lccc}
        \toprule
        
        \textbf{Model} & Training Method & Batch Size & Learning Rate \\
        
        \midrule
        
        \multirow{2}{*}{LLaVA-OV-0.5B} & DPO & 128 & 5e-7\\
                                     & SFT &   16  & 2e-6\\
        \hdashline\noalign{\vskip 0.5ex}
        \multirow{2}{*}{InternVL-2.5-MPO-1B} & DPO & 256 & 1e-6\\
                                     & SFT &   16  & 4e-5\\
        \hdashline\noalign{\vskip 0.5ex}
        \multirow{2}{*}{InternVL-2.5-MPO-4B} & DPO & 256 & 1e-6\\
                                     & SFT &   16  & 4e-5\\

        \bottomrule
    \end{tabular}
    }
    \end{adjustbox}
    \vspace{-2mm}
    
    \caption{Experimental details for different training approaches. All models are trained for one epoch for fair comparisons. }
    \label{tab:training_setting}
    \vspace{-5mm}
\end{table*}